\documentclass{article}

\usepackage{amsthm}
\usepackage{amsmath,amssymb}
\usepackage{amscd}
\usepackage{url}

\usepackage{comment}
\usepackage{babel}

\usepackage{array}   % for \newcolumntype macro
\newcolumntype{L}{>{$}l<{$}} % math-mode version of "l" column type

\newtheorem{definition}{Definition}
\newtheorem{example}[definition]{Example}

\newtheorem{proposition}[definition]{Proposition}

\begin{document}

%%\usepackage[colorlinks,urlcolor=blue]{hyperref}
%%\usepackage{comment}
%% The amssymb package provides various useful mathematical symbols
%%\usepackage{amssymb}
%% The amsthm package provides extended theorem environments
%% \usepackage{amsthm}
%%
%%\usepackage{graphicx}
%%\usepackage[latin1]{inputenc}
%%\usepackage{amssymb,amsmath,array}
%%\usepackage{eusflat2013}
%%\usepackage{lmodern} % Times fonts
%%\usepackage{color}

%% The lineno packages adds line numbers. Start line numbering with
%% \begin{linenumbers}, end it with \end{linenumbers}. Or switch it on
%% for the whole article with \linenumbers after \end{frontmatter}.
%% \usepackage{lineno}

%% natbib.sty is loaded by default. However, natbib options can be
%% provided with \biboptions{...} command. Following options are
%% valid:

%%   round  -  round parentheses are used (default)
%%   square -  square brackets are used   [option]
%%   curly  -  curly braces are used      {option}
%%   angle  -  angle brackets are used    <option>
%%   semicolon  -  multiple citations separated by semi-colon
%%   colon  - same as semicolon, an earlier confusion
%%   comma  -  separated by comma
%%   numbers-  selects numerical citations
%%   super  -  numerical citations as superscripts
%%   sort   -  sorts multiple citations according to order in ref. list
%%   sort&compress   -  like sort, but also compresses numerical citations
%%   compress - compresses without sorting
%%
%% \biboptions{comma,round}

% \biboptions{}

%%\journal{Nuclear Physics B}

%\begin{frontmatter}

\title{The transport problem for non-additive measures}
%\date{10 March 2022}
%%\author[a]{Vicen\c{c} Torra \\ %%}
\author{Vicen\c{c} Torra \\ %%}
%%
%%  \address[a]{
  Dept. Computing Sciences, Ume\aa~University, Sweden \\
  {\tt vtorra@cs.umu.se}}

\maketitle

\begin{abstract}
  Non-additive measures, also known as fuzzy measures, capacities, and monotonic games, are increasingly used in different fields. Applications have been built within computer science and artificial intelligence related to e.g. decision making, image processing, machine learning for both classification, and regression. Tools for measure identification have been built. In short, as non-additive measures are more general than additive ones (i.e., than probabilities), they have better modeling capabilities allowing to model situations and problems that cannot be modeled by the latter. See e.g. the application of non-additive measures and the Choquet integral to model both Ellsberg paradox and Allais paradox. 

  Because of that, there is an increasing need to analyze non-additive measures. The need for distances and similarities to compare them is no exception. Some work has been done for defining $f$-divergence for them. In this work we tackle the problem of defining the optimal transport problem for non-additive measures. Distances for pairs of probability distributions based on the optimal transport are extremely used in practical applications, and they are being studied extensively for their mathematical properties. We consider that it is necessary to provide appropriate definitions with a similar flavour, and that generalize the standard ones, for non-additive measures. 

  We provide definitions based on the M\"obius transform, but also based on the $(\max, +)$-transform that we consider that has some advantages. We will discuss in this paper the problems that arise to define the transport problem for non-additive measures, and discuss ways to solve them. In this paper we provide the definitions of the optimal transport problem, and prove some properties. 
\end{abstract}

%\begin{keyword}



%\end{keyword}

%\end{frontmatter}

% \linenumbers

%\setlength\paperheight{297mm}
%\setlength\paperwidth{210mm}

%%\thispagestyle{empty}

\section{Introduction}
The optimal transport~\cite{{ref:Villani.2003},{ref:Villani.2008},{ref:Santambrogio.2015},{ref:Bogachev.Kolesnikov.2012}} is a well known problem that has been studied from a theoretical perspective and it is currently extensively used in applications. The optimal transport problem was introduced by Kantorovich~\cite{ref:Kantorovich.1942} and is a generalization of Monge's optimal transport problem. Kantorovich's optimal problem can be seen as establishing a relationship -- an assignment -- between two probability spaces. This relationship is optimal with respect to an underlying cost function. The cost function combined with the optimal assignment can then be used to define a distance between the two probability distributions. It is the Wasserstein distance. There are a large number of applications~\cite{ref:Santambrogio.2015} of the optimal transport and the Wasserstein distance in statistics and machine learning. For example, the Wasserstein GANs~\cite{ref:Arjovsky.Chintala.Bottou.2017}.

Non-additive measures~\cite{{ref:Torra.Narukawa.Sugeno.2013}}, also known as fuzzy measures~\cite{{ref:Sugeno.1972},{ref:Sugeno.1974}} and monotonic games, generalize probabilities by replacing the additivity condition by a less restrictive one. More particularly, they just require measures to be monotonic with respect to set inclusion. Distorted probabilities~\cite{{ref:Chateauneuf.1996},{ref:Edwards.1953},{ref:Honda.2014},{ref:Honda.Nakano.Oakazaki.2002:Distortion}} are an example of these measures. Non-additive measures are used in economics, decision making, and artificial intelligence. They permit to model situations that cannot be modeled with probability measures. For example, both Ellsberg paradox and Allais paradox, which correspond to decision problems, cannot be represented with probabilities but they are representable (solved) using non-additive measures~\cite{ref:Gilboa.2009}. Similarly, some classification and regression problems~\cite{ref:Tehrani.et.al.2012} can be better solved with non-additive measures because they have additional degrees of freedom than additive ones (probabilities).

A fundamental difference between additive and non-additive measures is that the former consider the elements of the singletons independent while this is not so for non-additive ones. That is, for a non-additive measure $\mu$ on $X=\{x_1, x_2, \dots, x_n\}$ we may have $\mu(\{x_1, x_2\}) > \mu(\{x_1\}) + \mu(\{x_2\})$ if we have a positive interaction of $x_1$ and $x_2$, or we may have $\mu(\{x_1, x_2\}) < \mu(\{x_1\}) + \mu(\{x_2\})$ if we have a negative interaction of $x_1$ and $x_2$. For probabilities, only equality is possible. See e.g.~\cite{ref:Torra.Guillen.Santolino.2018:INFUS} for more concrete examples on what can be modeled with non-additive measures and cannot be modeled with probabilities. Then, in applications~\cite{{ref:Bardozzo.et.al.2021},{ref:Beliakov.2022},{ref:Marco-Detchart.et.al.2021},{ref:Pereira.Figueira.Marques.2020}}, the measures represent some background knowledge on the variables or attributes, and the non-additive integrals~\cite{{ref:Benvenuti.Mesiar.Vivona.2002},{ref:Mesiar.Mesiarova.2004},{ref:Torra.Narukawa.Sugeno.2013}} are used to aggregate or combine the data with respect to the measure. The two most used non-additive integrals are the Choquet~\cite{ref:Choquet.1953.54} and Sugeno~\cite{ref:Sugeno.1974} integrals, but there are several rich families of integrals (see e.g.~\cite{{ref:Mesiar.et.al.2016},{ref:Pereira-Dimuro.et.al.2020:CI.generalizations},{ref:Pereira-Dimuro.et.al.2020}}).

The more non-additive measures are used in applications, the more we need tools to compare and assess them. This naturally includes a need for similarities, distances, metrics, and divergences between pairs of measures. That is, we need a way to know when two measures are similar or not, and the extent of this similarity. Previous research on distances for non-additive measures exist. In particular, there are approaches to extend $f$-divergences through Radon-Nikodym-like derivatives. We have contributed~\cite{{ref:Torra.Narukawa.Sugeno.Carlson.2013},{ref:Torra.Narukawa.Sugeno.2016},{ref:Torra.Narukawa.Sugeno.2020}} to some of these results. See also the works by Agahi~\cite{{ref:Agahi.2020},{ref:Agahi.Yadollahzadeh.2021}}. The definition of $f$-divergences permits us to study Kullback–Leibler divergence, and define entropy, as well as the Principle of Minimum Discrimination and of Maximum Entropy~\cite{ref:Torra.2017:entropy}. This links with previous research on entropy for measures~\cite{ref:Honda.2014}. 

%%The optimal transport~\cite{{ref:Villani.2003},{ref:Villani.2008},{ref:Santambrogio.2015}} has been used very successfully for additive measures in all kind of applications. 

As the optimal transport problem has interesting mathematical properties and has been extensively used in applications, it seems a natural approach for non-additive ones as well. We study and formalize this problem in this paper. We can also solve it in practice for measures on a finite set defining an appropriate optimization problem, which is a linear optimization problem with linear constraints. Up to our knowledge no solution has been provided in the literature. This is so because of the difficulty to deal with the measure on non-singletons and to deal with positive and negative interactions on these sets. Let us briefly discuss some of these difficulties. Additional details and appropriate formalization will be given later in the paper. 

\begin{itemize}
\item Transport problems can be understood as the transfer of mass (i.e., probability) from one object to another one. In probabilities, transfer is between elements of a reference set. Then, for non-additive measures, we need to take into account, not only the transfer from singletons to singletons, but also from and to sets of arbitrary cardinality.
\item The mass associated to a non-singleton $A$ is not independent but, due to the monotonicity condition, naturally depends on the mass associated to singletons for $x \in A$ and, in general, to any subset $A'$ of the set $A$. Any definition of mass transfer needs to take this characteristic into account. In other words, it may be inappropriate to consider, in a solution, assigning mass associated to $\{x_1\}$ and $\{x_2\}$ to a set $A$ and assigning the mass of $\{x_1, x_2\}$ to a set $B$ such that $A \cap B = \emptyset$.
\item In a way, the total mass associated to two non-additive measures $\mu$ and $\nu$ is not necessarily the same even when $\mu(X)=\nu(X)=1$. For example, while in a probability, the total probability assigned to singletons is one, this is not necessarily true for non-additive measures. Note that it can be any number in the interval $[0,n]$, where $n$ is the cardinality of $X$. This adds an additional complexity into the process. This problem has some connections with the case of unbalanced transport problems studied in the optimal problem transport community. They are problems in which two measures do not have the same total mass. 
\end{itemize}

It is relevant to underline that for non-additive measures, the family of distances and similarities related to the $f$-divergence are based on the Choquet integral (i.e., an integral with respect to a non-additive measure). Their Radon-Nikodym-like derivative briefly mentioned above is defined taking into account such integral. In contrast, the definitions of the transport problem discussed in this paper, do not depend on this integral and is based on standard integration (i.e., addition in the discrete case).

\begin{itemize}
\item The transport problem considers costs and assignments, and the addition of their product to evaluate a solution. The integral is used when the domains are continuous. If we consider non-additive measures, an approach can consist on using non-additive integrals when computing this cost. Nevertheless, the Choquet integral does not satisfy, in general, Fubini's theorem. This approach is sketched in~\cite{ref:Gal.Niculescu.2019}. Nevertheless, it is difficult to solve this approach in practice, and the projection of the measures on the product space into each of the original spaces would need to be clearly defined. The results we provide in this work can help on this definition. 
\end{itemize}

The structure of the paper is as follows. In Section~\ref{sec:2} we review the concepts we will use in the rest of the paper. In Section~\ref{sec:3} we introduce our definitions of the transport problem for non-additive measures. We present our main results and discuss the relationship with the standard transport problem. The paper finishes with some research directions for future work.

\section{Preliminaries}
\label{sec:2}
We divide this section in two parts. One reviewing the transport problem and the other reviewing some definitions we need related to non-additive measures. In what follows, we consider measures on finite reference sets. 

\subsection{The transport problem}
Let us consider two additive measures $P$ and $Q$ on $X$ and $Y$, respectively, with probability distributions $p$ and $q$. Then, the transport problem~\cite{{ref:Villani.2003},{ref:Villani.2008},{ref:Santambrogio.2015}} consists on finding an assignment $\gamma$ from $p$ to $q$. The assignment $\gamma$ needs to have as marginals $p$ and $q$.

Then, given a cost function $c: X \times Y \rightarrow {\mathbb R}^+$, the optimal transport problem corresponds to the assignment that minimizes the total cost, where this total cost is defined by
$$\sum_{x \in X} \sum_{y \in Y} c(x,y) \gamma(x,y)$$
over the space of all possible assignments $\gamma$ with marginals $p$ and $q$.

It is usual to consider the transport problem for pairs of measures on the same reference set. That is, $X=Y$. This is a requirement when we define Wasserstein distance (or Kantorovich–Rubinstein metric) in terms of the optimal transport problem. More formally, the Wasserstein distance is defined as
\begin{equation}
  \label{eq:Wasserstein.probs}
  d(P,Q)=\inf_{\gamma \in \Gamma(P,Q)} \sum_{x \in X} \sum_{y \in Y} c(x,y) \gamma(x,y),
\end{equation}
where $c(x,y)=|x-y|$.

Note that other cost functions are also used. 

\subsection{Non-additive measures}
We begin this review defining a measure on a reference set $X$. We consider that this set is finite. Details on non-additive measures can be found in several reference works~\cite{{ref:Denneberg.1994},{ref:Torra.Narukawa.Sugeno.2013},{ref:Chateauneuf.Jaffray.1989}}. 

\begin{definition}
  Given a finite reference set $X$, a set function $\mu: 2^{X} \rightarrow [0,1]$ is a non-additive measure if:
  \begin{itemize}
  \item $\mu(\emptyset)=0$ (boundary condition)
  \item If $A \subseteq B$ then $\mu(A) \leq \mu(B)$ for $A,B \subseteq X$.
  \end{itemize}
\end{definition}

Observe that non-additive measures are also known as capacities, fuzzy measures, and monotonic games. 

If the measure is such that $\mu(X)=1$ we say that the measure is normalized.

There are alternative ways to represent non-additive measures. The M\"obius transform is one of them. We review its definition. We will also present the $(\max, +)$-transform. The latter transform has some connections with the generalizations~\cite{ref:Mesiar.1999} of $k$-order additive measures. See~\cite{ref:Torra.2023:max.plus.additive} for details.

\begin{definition}
  Let $\mu$ be a non-additive measure on $X$. Then, its M\"obius transform $\tau_{\mu}$ corresponds to:
  $$\tau_{\mu}(A) = \sum_{B \subseteq A} (-1)^{|A|-|B|}\mu(B).$$
\end{definition}

In this definition we use $\subseteq$ to denote the non-strict inclusion, and $\subset$ to denote the strict inclusion. 

Then, given a function $\tau_{\mu}$ that is the M\"obius transform of $\mu$, we have that for all $A \subseteq X$:
$$\mu(A)=\sum_{B \subseteq A}\tau_{\mu}(B).$$

In general, a set function $m$ on $X$ is a M\"obius transform of a monotone measure if
\begin{itemize}
\item (i) $\sum_{B \subseteq A}m(B) \geq 0$ for all $A \subseteq X$, and
\item (ii) for $A \subseteq A'$ it holds $\sum_{B \subseteq A} m(B) \leq \sum_{B \subseteq A'} m(B)$.
\end{itemize}
We can also add a condition for the normalization of the resulting measure. This naturally corresponds to require $\sum_{A \subseteq X} m(A)=1$.

A well known class of measures are the so-called belief functions. They are characterized by the fact that the M\"obius transform is always non-negative and adds to one (i.e., $m(A)\geq 0$) and $\sum_{A \subseteq X} m(A)=1$. In this case, the M\"obius transform is called a basic probability assignment (bpa). It is easy to see that for singletons $\{x_i\}$, the measure corresponds to the basic probability assignment or M\"obius transform of $\mu$. That is, i.e., $\mu(\{x_i\})=m(\{x_i\})$. 

We discuss now an alternative transform. 

\begin{definition}~\cite{ref:Torra.2023:max.plus.additive}
  Let $\mu$ be a non-additive measure on $X$. Then, we define the $(Max, +)$-transform as the set function $\tau_{\mu}: 2^X \rightarrow {\mathbb{R}}^+$ such that:
  \begin{equation}
    \label{eq:m.from.mu.max.plus}
    \tau_{\mu}(B) = \mu(B) - \max_{A \subset B} \mu(A)
  \end{equation}
\end{definition}

It can be seen from this definition that $\tau_{\mu}$ is always positive and that for an arbitrary normalized measure $\mu$ and any set $A \subseteq X$, the $(\max, +)$-transform $\tau_{\mu}(A)$ is at most one. That is, $\tau_{\mu}(A) \in [0,1]$ for all $A \subseteq X$. 

When $\mu$ is an additive measure, the following holds. We will use this result later to study the transport problem when the measures are additive. 

\begin{proposition}
  \label{prop:probability.distribution}
  Let $X$ be a reference set, and $\mu$ be a measure that is additive. Then, if $\tau_{\mu}$ is the $(\max,+)$-transform of $\mu$, the following holds:
  $$\tau_{\mu}(B) = \min_{x_i \in B} \mu(\{x_i\}).$$
\end{proposition}

\begin{proof}
  If $\mu$ is additive, then $\max_{A \subset B} \mu(A)$ will be achieved with the set $A_0$ containing $|A_0|=|A|-1$ elements $x_i$. These elements $x_i \in A$ are the ones with the largest values $\mu(\{x_i\})$. Therefore, the one that is missing in $A_0$ is the one $x_0$ with the smallest value $\mu(\{x_0\})$. Thus, $\mu(B) - \mu(A_0) = \mu(\{x_0\})=\min_{x_i \in B} \mu(\{x_i\})$. 
\end{proof}

We can also prove the following. 

\begin{proposition}~\cite{ref:Torra.2023:max.plus.additive}
  \label{prop:build.mu.from.m}
  Let $m$ be a set function over the set $X$ such that $m: 2^X \rightarrow [0,1]$, with $m(\emptyset)=0$. Then, the set function defined by 
  \begin{equation}
    \label{eq:mu.from.m.max.plus}
    \mu(B) = \left(\max_{A \subset B} \mu (A)\right) + m(B),
  \end{equation}
  is a non-additive measure and its $(Max, +)$-transform is $m$.   
\end{proposition}

Observe that, in general, for an arbitrary positive set function $m$ as above, the resulting measure is not necessarily normalized. For a given $\mu$, and its corresponding $(Max, +)$-transform $m$, we have that Proposition~\ref{prop:build.mu.from.m} returns $\mu$ as the measure associated to $m$. 

\section{Definitions}
\label{sec:3}
We will provide now alternative definitions for the transport problem. We begin using the basic probability assignment, which provides the definition that is closest to standard transport problems for probability distributions. We will show the limitations of this approach, and the difficulties to extend it to other types of non-additive measures. It will also provide a direction for other approaches. 

We will consider, in general, two non-additive measures $\mu$ and $\nu$, both defined on the same reference set $X$. The definitions we provide for the transport problem are easily generalized to the case of $\mu$ and $\nu$ defined on different reference sets $X$ and $Y$. 

\subsection{Beliefs functions and basic probability assignments}
We begin considering the transport problem for belief measures. That is, we will consider two belief measures $\mu$ and $\nu$ on the reference set $X$. As we have seen above, for belief measures, the M\"obius transform is called a basic probability assignment (bpa) and is a function $m$ such that $m(A)\geq 0$ for all $A$. As, in addition, $\sum_{A \subseteq X} m(A)=1$, basic probability assignments can be understood as probability distributions on the power set of $X$. Therefore, we can apply the definition of the optimal transport problem using as the reference set the power set of our original reference set. That is, $2^X$. This type of definition appears in~\cite{ref:Bronevich.Rozenberg.2021}.

\begin{definition}
  \label{def:OTp.bpa}
Let $\mu$ and $\nu$ be two belief functions with $\tau_{\mu}$ and $\tau_{\nu}$ be the corresponding M\"obius transforms. That is, $\tau_{\mu}$ and $\tau_{\nu}$ are basic probability assignments and, thus, they are positive and add to one. Then, given a cost function $c_b: 2^X \times 2^X \rightarrow {\mathbb{R}}^+$ we define the corresponding transport problem as follows.

Find the assignment $assg: 2^X \times 2^X \rightarrow [0,1]$ that minimizes the following objective function: 
\begin{equation}
  \label{eq:OTp.bpa}
  OF = \sum_{A \subseteq X} \sum_{B \subseteq X} c_b(A, B) assg(A, B)
\end{equation}

In the definition, $assg$ is a function $assg: 2^{X} \times 2^{X} \rightarrow [0,1]$, and the marginals of this function need to correspond to $\tau_{\mu}$ and $\tau_{\nu}$. In other words, $\tau_{\mu}(A)=\sum_{B \subseteq X} assg(A,B)$ and $\tau_{\nu}(B)=\sum_{A \subseteq X} assg(A,B)$.
\end{definition}

In this definition, $A$ or $B$ can be the empty set, but this does not play any role as the basic probability assignments associated to the empty sets are zero for both $\mu$ and $\nu$, and the assignment is non-negative.  

Let us consider some properties. The proof of the first one is trivial from the definition above. 

\begin{proposition}
Let $\mu$ and $\nu$ as above, and let $assg$ be as in the previous definition; then, $\sum_{A \subseteq X} \sum_{B \subseteq X} assg(A,B)=1$.
\end{proposition}

\begin{proposition}
  \label{prop:mu.bpa.equivalence}
  The following two conditions are equivalent
  \begin{itemize}
  \item $\sum_{B \subseteq X} assg(A,B) = \tau_{\mu}(A)$ for all $A \subseteq X$, and 
  \item $\mu^*(A)=\sum_{A^* \subseteq A} \sum_{B \subseteq X} assg(A^*, B) = \mu(A)$ for all $A \subseteq X$.
    \end{itemize}
  These conditions are given focusing on $\tau_{\mu}(A)$. A similar proposition can be proven when we consider $\tau_{\nu}(B)$ for all $B\subseteq X$. 
\end{proposition}

\begin{proof}
  To prove this proposition, we prove
  \begin{itemize}    
  \item If $\sum_{B \subseteq X} assg(A,B) = \tau_{\mu}(A)$ for all $A \subseteq X$,
    \\ then $\mu^*(A)=\sum_{A^* \subseteq A} \sum_{B \subseteq X} assg(A^*, B) = \mu(A)$ for all $A \subseteq X$, 
  \item If $\mu(A)=\sum_{A^* \subseteq A} \sum_{B \subseteq X} assg(A^*, B)$ for all $A \subseteq X$,
    \\ then $\sum_{B \subseteq X} assg(A,B) = \tau_{\mu}(A)$ for all $A \subseteq X$.
  \end{itemize}
  
  The first implication is trivial. Let us define $\mu^*(A)=\sum_{A^* \subseteq A} \tau_{\mu}(A^*)$ which is naturally equivalent to $\mu(A)$ and then replace $\tau_{\mu}(A^*)$ by its corresponding expression on the left. 

  The second implication can be proven by induction. First we consider it for a set of cardinality 1. That is, we consider $A=\{x_i\}$. In this case the only $A^*$ to consider in the summatory is also $A^*=\{x_i\}$. So, the expression in the proof corresponds to $\mu(\{x_i\})=\sum_{B \subseteq X} assg(\{x_i\}, B)$ which is naturally equivalent to $\tau_{\mu}(\{x_i\})$. Then, we consider in the induction hypothesis that the condition is true for sets of cardinality $n$ smaller than the cardinality of $X$ (i.e., $n <|X|$). Let us now prove that it is also true for a set of cardinality $n+1$. That is, we consider
  $\mu(A)=\sum_{A^* \subseteq A} \sum_{B \subseteq X} assg(A^*, B)$ and rewrite it distinguishing $A$ from the other $A' \neq A$ in $A^*$. That is,
  $\mu(A)=\sum_{A^* \subset A} \sum_{B \subseteq X} assg(A^*, B) + \sum_{B \subseteq X} assg(A, B)$
  By the induction hypothesis, we have that $\sum_{B \subseteq X} assg(A^*, B) = \tau_{\mu}(A^*)$ therefore, as $\mu(A)=\sum_{A^* \subseteq A} \tau_{\mu}(A^*)$, 
  $$\mu(A) = \sum_{A^* \subseteq A} \tau_{\mu}(A^*)=\sum_{A^* \subset A} \tau_{\mu}(A^*) + \sum_{B \subseteq X} assg(A, B).$$
  So, $\sum_{B \subseteq X} assg(A, B)$ should be $\tau_{\mu}(A)$, and the proposition is proven.
\end{proof}

This proposition shows that both conditions can be equivalently used in the transport problem to denote the marginals of the measure. That is, we can use to restrict the values of the marginals of the assignment, either the basic probability assignment or the measure itself. We prefer to use the constraint using the basic probability assignment, as we have written in Definition~\ref{def:OTp.bpa}. This means considering less terms in the constraint, and it is simpler to interpret. In this way, the constraint for each $A \subset X$ corresponds to a row in the matrix that represents the assignment for pairs $(A,B)$. 

\subsection{M\"obius transform}
When the measure is not a belief function, the M\"obius transform will contain some negative values. It is important to note that we not only need to deal with negative values, but the values can be arbitrarily large, for an appropriate number of elements in the reference set. Let us illustrate this with an example. 

\begin{example}
  \label{ex:Mobius.unbounded}
  Let $\mu$ be a non-additive measure defined as zero for all sets of cardinality smaller than $n$, and 1 for all sets of cardinality at least $n$. That is, $\mu(A)=1$ if and only if $|A|\geq n$ (and $\mu(A)=0$ otherwise). 

  Then, its M\"obius transform will be zero for all sets of cardinality smaller than $n$, will be one for all sets of cardinality $n$. In addition, for any set of cardinality $n+1$ there will be $n+1$ subsets with cardinality $n$ and, so, its M\"obius transform will be $\tau_{\mu}(A) = \mu(A) - (n+1) = 1 - (n+1)=-n$. 
\end{example}

The transport problem as given in Definition~\ref{def:OTp.bpa} is, in general, unsuitable for assignments that can be negative (i.e., $assgn: 2^X \times 2^X \rightarrow {\mathbb{R}}$). It is easy to see that for any cost function such that $c_b(A,A)=c_b(B,B)=0$ and $c_b(A,B)=c_b(B,A)=\kappa > 0$ and a feasible assignment $assg$, we can define a new feasible assignment $assgn' = assg+assgn^*$ with $assg^*(A,A)=assg^*(B,B)=+\alpha$ and $assg^*(A,B)=assg^*(B,A)=-\alpha$. This assignment will increase arbitrarily the objective function. Taking these considerations into account it is natural to consider the absolute value of the assignment. This results into the following objective function. This objective function replaces the one in Definition~\ref{def:OTp.bpa} (Equation~\ref{eq:OTp.bpa}). We use here $c_M$ to denote the cost function (to distinguish it from the one for basic probability assignments $c_b$. 

\begin{equation}
  \label{eq:OTp.mobius}
  OF = \sum_{\emptyset \subset A \subseteq X} \sum_{\emptyset \subset B \subseteq X} c_M(A, B) |assg(A, B)|
\end{equation}

The assignment $assg$ needs to satisfy the marginals, either positive or negative, about the M\"obius transform. Observe that here the equivalences on the expressions in Proposition~\ref{prop:mu.bpa.equivalence} also apply.

We have shown in Example~\ref{ex:Mobius.unbounded} that we can have a M\"obius transform with very large values. We will now consider two measures similar to Example~\ref{ex:Mobius.unbounded} and show its effect in the value of the objective function. The example will illustrate that, in a way, the mass is counted multiple times in the cost function. 

\begin{example}
  Let $\mu$ and $\nu$ be two non-additive measures defined on a reference set $X=\{x_1, \dots, x_n, x_{n+1}, \dots, x_{3n}\}$. Let $A_0 = \{x_1, \dots, x_n, x_{n+1}\}$, and let $B_0=\{x_{n+2}, \dots, x_{2n+2}\}$.

  Let us define $\mu$ as follows. Let $\mu(A)=1$ for all subsets of $A_0$ with cardinality $n$. Let $\mu(A)=0$ for all subsets of $A_0$ with cardinality smaller than $n$. Let $A_i=A_0 \setminus \{x_i\}$ for $i=1, \dots, n+1$. Therefore, for these sets $\mu(A_i)=1$, as they have cardinality $n$. In addition, all supersets of these sets will also have measure one.

  Similarly, let $\nu(B)=1$ for all subsets of $B_0$ with cardinality $n$, and $\nu(B)=0$ for all subsets of $B_0$ with cardinality smaller than $n$. Let $B_i=B_0 \setminus  \{x_{n+1+i}\}$, so $\nu(B_i)=1$. Similarly, all supersets of these sets will also have measure one. 

  Let us now consider a cost function. For simplicity, the cost function is $c_M(A,B)=1$ for $A \neq B$ and $c_M(A,B)=0$ for $A = B$. 

  Then, a natural assignment is to assign $assg(A_i, B_i)=1$ for $i=1, \dots,n$. Nevertheless, we also need to assign the mass for the other sets with non zero M\"obius transform. Note that the M\"obius transform for both $A_0$ and $B_0$ will be $-n$. So, we need e.g. $assg(A_0,B_0)=-n$. 
\end{example}

This example illustrates that the definition of the problem in terms of the M\"obius transform causes the mass of some sets appear more than once in the cost. I.e., the value of $-n$ associated to $A_0$ mainly corresponds to the mass of all $A_i$ which we also need to assign. We have assigned it to $B_i$ with $assg(A_i, B_i)$. In other words, we have transferred one unit from $A_1$ to $B_1$ with $assg(A_1, B_1)=1$, and the same from $A_i$ to $B_i$, and then we need to transfer this mass again through the assignment $assg(A_0,B_0)=-n$.

We have focused on large negative values. It is important to note that we can also have very large positive values for the M\"obius transform. The following example illustrates this case. 

\begin{example}
  Consider a measure $\mu$ similar to the one above. In this case, we use $A_0 = \{x_1, \dots, x_n, x_{n+1}, x_{n+2}\}$. Let us denote by $A^{-}$ the subsets of $A_0$ of cardinality $n+1$, and $A^{--}$ the subsets of $A_0$ of cardinality $n$. Then, define $\mu(A)=1$ for all subsets $A^{-}$ and $A^{--}$ (and, naturally, also for all supersets of these sets). Let $\mu(A)=0$ for all subsets of $A_0$ with cardinality smaller than $n$. 

Then, using the discussion above we have that $\tau_{\mu}(A)=1$ for all sets $A \in A^{--}$, and $\tau_{\mu}(A)=-n$ for all sets $A \in A^{-}$. In addition, for smaller sets we have $\tau_{\mu}(A)=0$. Then, for $A_0$ we have
\begin{eqnarray}
  \mu(A_0)=1
  &=&\tau_{\mu}(A_0) + \sum_{A\in A^{-}} \tau_{\mu}(A) + \sum_{A\in A^{--}} \tau_{\mu}(A) \nonumber \\
  &=& \tau_{\mu}(A_0) + (n+2)(-n) + \frac{(n+2)(n+1)}{2} \cdot 1. \nonumber
\end{eqnarray}
Therefore, $\tau_{\mu}(A_0)=(n^2+n)/2$. 
\end{example}

The formulation of the optimal transport using the $(\max, +)$-transform mitigates this problem. This is discussed in the next section. As we will see, this is at the cost of having unassigned mass. 

\subsection{$(max, +)$-transform} 
The two definitions above focus on the M\"obius transform. We now consider the $(max, +)$-transform. 

A first thought is to use the same approach as in Definition~\ref{def:OTp.bpa} but using now the $(\max, +)$-transform. That is, use Equation~\ref{eq:OTp.bpa} and the marginals that correspond to $\tau_{\mu}(A)$ and $\tau_{\nu}(B)$ where these expressions correspond to: 
$$\tau_{\mu}(A)=\sum_{\emptyset \subset B^* \subseteq X} assg(A, B^*)$$
$$\tau_{\nu}(B)=\sum_{\emptyset \subset A^* \subseteq X} assg(A^*,B)$$

Unfortunately, this approach does not always lead to a solution. That is, there are pairs of measures $\mu$ and $\nu$ for which there is no feasible assignment. This is illustrated in the following example.

\begin{example}
  \label{ex:mu.additive.nu.non.additive}
  Let $\mu$ and $\nu$ be two measures on $X = \{x_1, x_2, x_3\}$. Let $\mu$ be additive with measures in the singletons equal to 0.2, 0.3, and 0.5. Let $\nu(\{x_1\})=0.2$, $\nu(\{x_1, x_2\})=0.2$, $\nu(\{x_1, x_3\})=0.2$, $\nu(X)=1$ and $\nu(A)=0$ for all the other sets. Then, there is no assignment that is consistent with the marginals of $\tau_{\mu}$ and $\tau_{\nu}$. Table~\ref{ex:0.2.additive} displays the measures for all subsets of $X$, the values of $\tau_{\mu}$ and $\tau_{\nu}$, and an assignment that is consistent with all marginals except the one for $\tau_{\nu}(X)$. 
\end{example}

\begin{table}[ht]
  \scriptsize 
  %\begin{tabular}{ L | $l$ | $l$ || $l$ $l$ $l$ | $l$ $l$ $l$ | $l$ |}
  \begin{tabular}{ L | L | L || L L L  | L L L | L |}
    \nu(B)& \tau_{\nu} & set         &        &      &    &      &    &     &     \\
    \hline
    \hline
    1   &  0.8 & \{x_1, x_2, x_3\} & 0     & 0.3  & 0.5 & 0.1 & 0.1 & 0.4 & 0.1 \\ 
    \hline 
    0   &  0        & \{x_2, x_3\} & 0     & 0    & 0   & 0    & 0  & 0   & 0   \\
    0.2 &  0        & \{x_1, x_3\} & 0     & 0    & 0   & 0    & 0  & 0   & 0   \\
    0.2 &  0        & \{x_1, x_2\} & 0     & 0    & 0   & 0    & 0  & 0   & 0   \\
    \hline
    0   &  0        & \{x_3\}      & 0     & 0    & 0   & 0    & 0  & 0   & 0   \\
    0   &  0        & \{x_2\}      & 0     & 0    & 0   & 0    & 0  & 0   & 0   \\ 
    0.2 &  0.2      & \{x_1\}      & 0.2   & 0    & 0   & 0    & 0  & 0   & 0   \\
    \hline
    \hline
    set  &          &              & \{x_1\} &\{x_2\} &\{x_3\} & \{x_1, x_2\} &\{x_1, x_3\} &\{x_2, x_3\} &\{x_1, x_2, x_3\} \\ 
    \tau_{\mu} & --  &              & 0.2   & 0.3  & 0.5 & 0.1 & 0.1 & 0.4 & 0.1 \\ 
    \mu(A)&           &            & 0.2   & 0.3  & 0.5 & 0.4 & 0.6 & 0.9 & 1.0 \\
\end{tabular}
\caption{Measures $\mu$ and $\nu$ of Example~\ref{ex:mu.additive.nu.non.additive}, their $(\max,+)$-transforms $\tau_{\mu}$ and $\tau_{\nu}$, and a non-feasible assignment. \label{ex:0.2.additive}}
\end{table}

In order that there is a possible assignment for any pair of measures, we need to relax the problem. The relaxation is to allow lack of assignment for any set and measure. More formally, for a measure $\mu$ and the set $A$, we consider the value $lack_{\mu}(A)$, and for $\nu$ and the set $B$, we consider the value $lack_{\nu}(B)$. Then, an assignment $assg$ needs to be consistent with the measures given $lack_{\mu}$ and $lack_{\nu}$. In other words, the following two equations need to hold: 

$$\mu(A)=\max_{A' \subset A} \mu(A') + \sum_{\emptyset \subset B^* \subseteq X} assg(A,B^*) + lack_{\mu}(A)$$
$$\nu(B)=\max_{B' \subset B} \nu(B') + \sum_{\emptyset \subset A^* \subseteq X} assg(A^*,B) + lack_{\nu}(B)$$

It is easy to see that these equations imply that the two right-most terms need to correspond to the $(\max,+)$-transforms $\tau_{\mu}(A)$ and $\tau_{\nu}(B)$, respectively. We will use these equivalent equations using the transform in Definition~\ref{def:transport.problem.mu.lack}. Table~\ref{ex:0.2.additive.with.lack} gives a possible solution adapting the previous non-feasible assignment with the $lack$ values. More particularly, we define $lack_{\nu}(X)=-0.7$ to make the equality possible. Nevertheless, this solution has a negative value. We present a formalization below that forces all assignments to be non-negative. We will prove that there are feasible solutions in this case. 

%\begin{tabular}{ L | $l$ | $l$ || $l$ $l$ $l$ | $l$ $l$ $l$ | $l$ |}
\begin{table}[ht]
  \scriptsize
  \begin{tabular}{ L | L | L || L || L L L  | L L L | L |}
    \nu(B)& \tau_{\nu} & set   & lack_{\nu} &        &      &    &      &    &     &      \\
    \hline
    \hline
    1   &  0.8    &\{x_1, x_2, x_3\}&-0.7  & 0    & 0.3  & 0.5 & 0.1 & 0.1 & 0.4& 0.1 \\ 
    \hline 
    0   &  0        & \{x_2, x_3\} & 0    & 0    & 0    & 0   & 0   & 0   & 0   & 0   \\
    0.2 &  0        & \{x_1, x_3\} & 0    & 0    & 0    & 0   & 0   & 0   & 0   & 0   \\
    0.2 &  0        & \{x_1, x_2\} & 0    & 0    & 0    & 0   & 0   & 0   & 0   & 0   \\
    \hline
    0   &  0        & \{x_3\}      & 0    & 0    & 0    & 0   & 0   & 0   & 0   & 0   \\
    0   &  0        & \{x_2\}      & 0    & 0    & 0    & 0   & 0   & 0   & 0   & 0   \\ 
    0.2 &  0.2      & \{x_1\}      & 0    & 0.2  & 0    & 0   & 0   & 0   & 0   & 0   \\
    \hline
    \hline
    lack_{\mu} &     &              & 0   & 0   & 0    & 0   & 0   & 0   & 0   & 0   \\ 
\hline
    set  &          &              & --   & \{x_1\} &\{x_2\} &\{x_3\} & \{x_1, x_2\} &\{x_1, x_3\} &\{x_2, x_3\} &\{x_1, x_2, x_3\} \\ 
    \tau_{\mu} & --  &              & --   & 0.2   & 0.3  & 0.5 & 0.1 & 0.1 & 0.4 & 0.1 \\ 
    \mu(A)&         &              & --   & 0.2   & 0.3  & 0.5 & 0.4 & 0.6 & 0.9 & 1.0 \\
  \end{tabular}
  \caption{Measures $\mu$ and $\nu$ of Example~\ref{ex:mu.additive.nu.non.additive}, their $(\max,+)$-transforms $\tau_{\mu}$ and $\tau_{\nu}$, and an assignment with $lack_{\mu}$ and $lack_{\nu}$. This example does not conform with Definition~\ref{def:transport.problem.mu.lack} because the assignment to $lack_{\nu}$ includes a negative value. \label{ex:0.2.additive.with.lack}}
\end{table}

We formalize the optimal transport problem considering three cost functions. We consider three cost functions. One associated to $assg$, called call $c_{a}$, and one associated to each measure, called $c_{\mu}$ and $c_{\nu}$. These latter cost functions are naturally associated to $lack_{\mu}$ and $lack_{\nu}$, respectively.

\begin{definition}
  \label{def:transport.problem.mu.lack}
  Let $\mu$ and $\nu$ be two non-additive measures on the reference set $X$ with $(max, +)$-transform $\tau_{\mu}$ and $\tau_{\nu}$, respectively. Then, the transport problem between $\mu$ and $\nu$ corresponds to find the functions $assg: (2^X\setminus \emptyset) \times (2^X \setminus \emptyset) \rightarrow [0,1]$, $lack_{\mu}: (2^X\setminus \emptyset) \rightarrow [0,1]$, and $lack_{\nu}: (2^X\setminus \emptyset) \rightarrow [0,1]$ that satisfy
  $$\tau_{\mu}(A)=\sum_{\emptyset \subset B^* \subseteq X} assg(A,B^*) + lack_{\mu}(A)$$
  $$\tau_{\nu}(B)=\sum_{\emptyset \subset A^* \subseteq X} assg(A^*,B) + lack_{\nu}(B)$$
  We will denote the solution of such problem by the tuple $(assg, lack_{\mu}, lack_{\nu})$.

  Then, given cost functions $c_{a}: 2^X \times 2^X \rightarrow [0,1]$, $c_{\mu}: 2^X \rightarrow [0,1]$, and $c_{\nu}: 2^X \rightarrow [0,1]$, the cost of $(assg, lack_{\mu}, lack_{\nu})$ is: 
  $$\sum_{\emptyset \subset A \subseteq X}\sum_{\emptyset \subset B \subseteq X} c_{a}(A,B) assg(A,B) 
  + \sum_{A \subseteq X} c_{\mu}(A) lack_{\mu}(A)
  + \sum_{B \subseteq X} c_{\nu}(B) lack_{\nu}(B).$$
\end{definition}

It is now possible to prove that for any pair of measures, there is at least an assignment that solves this transport problem. 

\begin{proposition}
  Let $\mu$ and $\nu$ be two non-additive measures on the reference set $X$ with $(max, +)$-transform $\tau_{\mu}$ and $\tau_{\nu}$, respectively. Then, there exist an assignment that solves the problem stated in Definition~\ref{def:transport.problem.mu.lack}.
\end{proposition}

\begin{proof}
  Consider the assignment $lack_{\mu}(A)=\tau_{\mu}(A)$ and $lack_{\nu}(B)=\tau_{\nu}(B)$, and $assg(A,B)=0$ for all other sets. This assignment satisfies the requirements in Definition~\ref{def:transport.problem.mu.lack}.
\end{proof}

The following example gives an example of solution for Example~\ref{ex:mu.additive.nu.non.additive}.

\begin{example}
  Let us reconsider the measures in Example~\ref{ex:mu.additive.nu.non.additive}. Table~\ref{ex:0.2.additive.with.lack.feasible} shows a feasible solution for the transport problem according to Definition~\ref{def:transport.problem.mu.lack}.
\end{example}

%\begin{tabular}{ L | $l$ | $l$ || $l$ $l$ $l$ | $l$ $l$ $l$ | $l$ |}
\begin{table}[ht]
  \scriptsize
  \begin{tabular}{ L | L | L || L || L L L  | L L L | L |}
    \nu(B)& \tau_{\nu} & set   & lack_{\nu} &        &      &    &      &    &     &      \\
    \hline
    \hline
    1   &  0.8    &\{x_1, x_2, x_3\}& 0   & 0    & 0.3  & 0.5 & 0   & 0   & 0   & 0   \\ 
    \hline 
    0   &  0        & \{x_2, x_3\} & 0    & 0    & 0    & 0   & 0   & 0   & 0   & 0   \\
    0.2 &  0        & \{x_1, x_3\} & 0    & 0    & 0    & 0   & 0   & 0   & 0   & 0   \\
    0.2 &  0        & \{x_1, x_2\} & 0    & 0    & 0    & 0   & 0   & 0   & 0   & 0   \\
    \hline
    0   &  0        & \{x_3\}      & 0    & 0    & 0    & 0   & 0   & 0   & 0   & 0   \\
    0   &  0        & \{x_2\}      & 0    & 0    & 0    & 0   & 0   & 0   & 0   & 0   \\ 
    0.2 &  0.2      & \{x_1\}      & 0    & 0.2  & 0    & 0   & 0   & 0   & 0   & 0   \\
    \hline
    \hline
    lack_{\mu} &     &              & --   & 0   & 0    & 0   & 0.1 & 0.1 & 0.4& 0.1 \\ 
\hline
    set  &          &              & --   & \{x_1\} &\{x_2\} &\{x_3\} & \{x_1, x_2\} &\{x_1, x_3\} &\{x_2, x_3\} &\{x_1, x_2, x_3\} \\ 
    \tau_{\mu} & --  &              & --   & 0.2   & 0.3  & 0.5 & 0.1 & 0.1 & 0.4 & 0.1 \\ 
    \mu(A)&         &              & --   & 0.2   & 0.3  & 0.5 & 0.4 & 0.6 & 0.9 & 1.0 \\
  \end{tabular}
  \caption{Measures $\mu$ and $\nu$ of Example~\ref{ex:mu.additive.nu.non.additive}, their $(\max,+)$-transforms $\tau_{\mu}$ and $\tau_{\nu}$, and a feasible assignment with $lack_{\mu}$ and $lack_{\nu}$ according to Definition~\ref{def:transport.problem.mu.lack}. \label{ex:0.2.additive.with.lack.feasible}}
\end{table}

We have defined the problem in a way that $assg$ is not defined for empty sets. So, we can revisit the definition and express $lack_{\mu}(A)$ and $lack_{\nu}(B)$ as equal to $assg(A, \emptyset)$ and $assg(\emptyset, B)$, respectively. We can also proceed in the same way with the cost function extending it to any subset of $X$ including the empty set. The cost of an assignment to the empty set is the cost associated to $lack_{\mu}$ and $lack_{\nu}$. Using this approach, we can express the transport problem as follows. We also define below the optimal transport which is, of course, a solution with a minimum cost. 

\begin{definition}
  \label{def:transport.problem.mu.no.lack}
  Let $\mu$ and $\nu$ be non-additive measures on $X$, with $(max, +)$-transforms $\tau_{\mu}$ and $\tau_{\nu}$, respectively. Then, the transport problem between $\mu$ and $\nu$ is a function $assg: 2^X \times 2^X \rightarrow [0,1]$ that satisfies
  $$assg(\emptyset, \emptyset)=0$$
  $$\tau_{\mu}(A)= \sum_{B^* \subseteq X} assg(A,B^*)~\textrm{for all $A \neq \emptyset$}$$
  $$\tau_{\nu}(B)= \sum_{A^* \subseteq X} assg(A^*,B)~\textrm{for all $B \neq \emptyset$}$$

  Then, given the cost function $c_{a}: 2^X \times 2^X \rightarrow [0,1]$, the cost of the assignment $assg$ is: 
  $$cost(c_{a}, assg) =
  \sum_{A \subseteq X}\sum_{B \subseteq X} c_{a}(A,B) assg(A,B).$$
\end{definition}

\begin{definition}
  Let $\mu$, $\nu$, $\tau_{\mu}$, $\tau_{\nu}$, and $assg$ as in Definition~\ref{def:transport.problem.mu.no.lack}. Then, the optimal problem is to find an assignment $assg$ that minimizes $cost(c_{a},assg)$. 
\end{definition}

  \begin{table}[ht]
    \scriptsize
    \begin{tabular}{ L | L | L || L | L L L  | L L L | L |}
      \nu(B)& \tau_{\nu} & set       &     &       &     &    &     &     &     &     \\
      \hline
      \hline
      1   &  0.2 & \{x_1, x_2, x_3\}& 0    & 0     & 0   & 0      & 0           & 0          & 0           & 0.2 \\
      \hline 
      0.8 &  0.2     & \{x_2, x_3\} & 0    & 0     & 0   & 0      & 0           & 0          & 0.2         & 0  \\
      0.8 &  0.2     & \{x_1, x_3\} & 0    & 0     & 0   & 0      & 0           &0.2         & 0           & 0  \\
      0.4 &  0.2     & \{x_1, x_2\} & 0    & 0     & 0   & 0      & 0.2         &0           & 0           & 0  \\
      \hline
      0.6 &  0.6     & \{x_3\}      & 0    & 0     & 0.1 & 0.5    & 0           & 0          & 0           & 0  \\
      0.2 &  0.2     & \{x_2\}      & 0    & 0     & 0.2 & 0      & 0           & 0          & 0           & 0 \\ 
      0.2 &  0.2     & \{x_1\}      & 0    & 0.2   & 0   & 0      & 0           & 0          & 0           & 0 \\
      \hline
      0   &  0       & \emptyset    & 0    &  0    & 0   & 0      & 0           & 0          & 0.1         & 0\\ 
      \hline
      \hline
      set  &          &       & \emptyset &\{x_1\}&\{x_2\}&\{x_3\}& \{x_1, x_2\} &\{x_1, x_3\} &\{x_2, x_3\} &\{x_1, x_2, x_3\} \\ 
      \tau_{\mu} & --  &             & 0.0  & 0.2   & 0.3 & 0.5    & 0.2         & 0.2        & 0.3         & 0.2              \\
      \mu(A)&         &             & 0.0  & 0.2   & 0.3 & 0.5    & 0.5         & 0.7        & 0.8         & 1.0              \\
    \end{tabular}
    \caption{Additive case. This is not the only assignment. An alternative is $assg(\{x_2\},\{x_3\})=0.1$ instead of $assg(\{x_2\},\{x_2,x_3\})=0.1$ (and then replace  $assg(\{x_1,x_2,x_3\},\{x_3\})=0.1$ by $assg(\{x_1,x_2,x_3\},\{x_2,x_3\})=0.1$).
      \label{tab:additive.mu}
    }
  \end{table}

Let us consider an example in which two additive measures are involved. 

\begin{example}
  Let $\mu$ and $\nu$ be two additive measures on $X$. The measure $\mu$ is defined by $\mu(\{x_1\})=0.2$, $\mu(\{x_2\})=0.3$, and $\mu(\{x_3\})=0.5$. The measure $\nu$ is defined by $\nu(\{x_1\})=0.2$, $\nu(\{x_2\})=0.2$, and $\nu(\{x_3\})=0.6$. Table~\ref{tab:additive.mu} includes the measures as well as the $(\max, +)$-transforms $\tau_{\mu}$ and $\tau_{\nu}$. A feasible assignment $assg$ is also included. 
\end{example}

\section{Results}
In this section we provide some results related to our definitions. In particular, we show that our definitions generalize standard optimal transport problems. We also discuss how to solve the problems defined. For the sake of generality, we will consider pairs of measures $\mu$ and $\nu$ on reference sets $X$ and $Y$. 

\subsection{Optimal transports as proper generalizations}
One can ask how a solution for this problem in the setting of non-additive measures relates to the one we would obtain with the optimal transport problem in the classical probabilistic setting. This question can be stated in the following terms.

Let $\mu$ and $\nu$ be additive measures, is the optimal assignment according to the problems defined in this section be the optimal assignment using the standard definition?

First, observe that one optimal assignment for probabilities is defined considering pairs $(x,y)$ from $(X,Y)$. In contrast, optimal assignments for non-additive measures are defined considering pairs $(A,B)$ that are subsets of $2^X$ and $2^Y$. Because of that, we consider the equivalence of the latter when restricted to the pairs considered by the former.

\begin{definition}
  Let $a$ be an assignment for pairs $(x,y)$ where $x \in X$ and $y \in Y$ and let $assg$ be an assignment for pairs $(A,B)$ where $A \subseteq X$ and $B\subseteq Y$. Then, we define $a \prec assg$ if and only if $a(x,y) = assg(\{x\},\{y\})$ for all $x \in X$ and $y \in Y$. 
\end{definition}

Then, we can prove the following for the optimal transport problem based on basic probability assignments based on Definition~\ref{def:OTp.bpa}. 

\begin{proposition}
  \label{prop:bpa.assignment.is.p.assignment}
  Let $\mu$ and $\nu$ be two probability measures on finite reference sets $X$ and $Y$, let $c$ be the cost function of the optimal transport for probability measures, and let $\kappa$ be an arbitrary value such that $\kappa > \max c(x,y)$. Then, let us define $c_b(\{x\},\{y\})=c(x,y)$ for all $x \in X$ and $y \in Y$, and $c_b(A,B)=\kappa$ for all other pairs $(A,B)$. Finally, let $a$ be the optimal transport for probability distributions $\mu$ and $\nu$ for $c$, and let $assg$ be the optimal transport according to Definition~\ref{def:OTp.bpa}.  

  Then, the optimal assignments $a$ and $assg$ are such that $a \prec assg$, and the values of the objective functions of these two problems are equal. 
\end{proposition}

\begin{proof}
  It is easy to see that the basic probability assignment $\tau_{\mu}$ and $\tau_{\nu}$ for both measures $\mu$ and $\nu$ will have zero value in all elements that are not singletons. So, $assg$ will only take values on singletons. As $c$ is equal to $c_b$ on these sets, the proposition is proven. 
\end{proof}

Similarly, we can prove the following with respect to the optimal transport problem for arbitrary measures in terms of M\"obius transforms. That is, using the objective function in Equation~\ref{eq:OTp.mobius}. 

\begin{proposition}
  Let $\mu$, $\nu$, $c$ be as in Proposition~\ref{prop:bpa.assignment.is.p.assignment}. Let $c_M$ be a cost function defined from $c$ using the approach for $c_b$ in Proposition~\ref{prop:bpa.assignment.is.p.assignment}. Let $a$ the optimal transport for $\mu$ and $\nu$ using $c$, and $assg$ be the optimal transport obtained using Equation~\ref{eq:OTp.mobius} using $c_M$. Then, $a \prec assg$, and the values of the objective functions of these two problems are equal. 
\end{proposition}

\begin{proof}
  Note that while $assg$ can take negative values, the objective function takes the absolute value of these assignments. Therefore, any solution with smaller assignment on the pairs $(x,y)$ will need to have larger values for non-singletons and, thus, the objective function will be larger.
\end{proof}

Let us now consider the optimal transport with respect to the $(\max, +)$-transform. While for any probability measure the basic probability assignment and, in general, the M\"obius transform is zero for non-singletons, this is not the case for the $(\max, +)$-transform. Nevertheless, we can also obtain a similar theorem. 

\begin{proposition}
  \label{prop:optimal.assignment.equal.probs.max.plus.transform}
  Let $\mu$ and $\nu$ be two probability measures on finite reference sets $X$ and $Y$, let $c$ be the cost function of the optimal transport problem for probability measures. Let $\kappa$ be an arbitrary value such that $\kappa > \max c(x,y)$. 

  Then, let us define a cost function $c_a$ as follows: $c_a(\{x\},\{y\})=c(x,y)$ for all $x \in X$ and $y \in Y$, $c_a(\{x\},B)=c_a(A,\{y\})=\kappa$ for non-singletons $A \subseteq X$ and $B \subseteq X$, and $c_a(A,B)=0$ for all other pairs $(A,B)$. Let $a$ be the optimal transport for probability distributions for $\mu$ and $\nu$ using $c$, and let $assg$ be the optimal transport according to Definition~\ref{def:transport.problem.mu.no.lack} using this cost function $c_a$. 

  Then, the optimal assignments $a$ and $assg$ are such that $a \prec assg$, and the objective functions of the two problems are the same. 
\end{proposition}

\begin{proof}
  As $\mu$ and $\nu$ are probability measures, the $(\max, +)$-transforms on the singletons will correspond to their probabilities. As $c=c_a$ for the singletons, the optimal assignment $assg$ for the singletons is the optimal assignment $a$. Observe that as $c_a(\{x\},B)$ and $c_a(A,\{y\})$ is $\kappa$ for non-singletons (or for $A=\emptyset$ and $B=\emptyset$), no other assignment $assg$ for these sets can lead to a lower value. Then, there will some arbitrary assignments for the pairs $(A,B)$ when both are non-singletons (including $A=\emptyset$ and $B=\emptyset$).

  The contribution of singletons to the objective function will be equivalent to the objective function of $a$. The contribution of non-singletons to the objective function will be zero. Therefore, the proposition is proven. 
\end{proof}

\begin{table}[ht]
  \scriptsize
  \begin{tabular}{ L | L | L || L | L L L  | L L L | L |}
    \nu(B)& \tau_{\nu} & set       &     &       &     &    &     &     &     &     \\
    \hline
    \hline
    1   &  0.2 & \{y_1, y_2, y_3\}& 0.2    & 0   & 0   & 0      & 0           & 0          & 0           & 0 \\
    \hline 
    0.8 &  0.2     & \{y_2, y_3\} & 0.2    & 0   & 0   & 0      & 0           & 0           & 0           & 0   \\
    0.8 &  0.2     & \{y_1, y_3\} & 0.2    & 0   & 0   & 0      & 0           &0          &  0          &  0  \\
    0.4 &  0.2     & \{y_1, y_2\} & 0.2    & 0   & 0   & 0      & 0           &0           & 0           &  0  \\
    \hline
    0.6 &  0.6     & \{y_3\}      & 0.6    & 0   & 0   & 0      & 0           & 0          & 0           & 0  \\
    0.2 &  0.2     & \{y_2\}      & 0.2    & 0   & 0   & 0      & 0           & 0          & 0           & 0  \\ 
    0.2 &  0.2     & \{y_1\}      & 0.2    & 0   & 0   & 0      & 0           & 0          & 0           & 0  \\
    \hline
    0   &  0       & \emptyset    & 0    & 0.2   & 0.3 & 0.5    & 0.2         & 0.2        & 0           & 0                \\
    \hline
    \hline
    set  &          &       & \emptyset &\{x_1\}&\{x_2\}&\{x_3\}& \{x_1, x_2\} &\{x_1, x_3\} &\{x_2, x_3\} &\{x_1, x_2, x_3\} \\ 
    \tau_{\mu} & --  &             & 0.0  & 0.2   & 0.3 & 0.5    & 0.2         & 0.2        & 0.3         & 0.2              \\
    \mu(A)&         &             & 0.0  & 0.2   & 0.3 & 0.5    & 0.5         & 0.7        & 0.8         & 1.0              \\
  \end{tabular}
  \caption{Additive case. Mass assigned to empty sets. \label{tab:additive.mu.to.emptyset}    }
\end{table}

The properties in this section mean that for any cost function $c$ and probability measures $\mu$ and $\nu$, we have a cost function $c_a$ such that the optimal transport problem for $c$ corresponds to the optimal transport problem using basic probability assignments for belief functions, M\"obius transform for arbitrary non-additive measures, and $(\max, +)$-transform also for an arbitrary non-additive measure.

The propositions establish that the assignment on the singletons will be the same. Nevertheless, nothing is said on the assignment for the other sets. Observe that for both basic probability assignments and M\"obius transforms the mass for non-singletons will be zero, so, the optimal assignment for our problems will be also zero for these sets. In contrast, the $(\max,+)$-transform is not zero for the non-singletons and, therefore, the assignment will neither be zero for them.

\subsection{Some additional properties}
The optimal transport problem according to Definition~\ref{def:transport.problem.mu.no.lack} has an interesting property. Given two measures, it is possible to {\em transfer} all the mass from sets $\emptyset \neq A \subseteq X$ to the empty set (i.e., what corresponds to the lack variable $lack_{\mu}(A)$) and the same for all sets $\emptyset \neq B \subseteq Y$. This would produce the solution described in Table~\ref{tab:additive.mu.to.emptyset}. To avoid this type of transfer we need that the cost from a set $A$ to the emptyset, and from emptyset to a set $B$ should be larger than a direct assignment from $A$ to $B$. That is,
$$c_a(A, \emptyset)+c_a(\emptyset,B) \geq c_a(A,B).$$

In Proposition~\ref{prop:optimal.assignment.equal.probs.max.plus.transform} we have given a cost function $c_a$ that provides a solution that is compatible with the optimal transport problem for probabilities. We can have other assignments that have the same property. Given that $c(x, y)<\infty$, we can use a cost function $c_a$ satisfying the following conditions:

\begin{itemize}
\item $c_a(\{x\}, \{y\}) = c(x,y)$,
\item $c_a(\{x\}, \emptyset) = \kappa^+ > max_{y_i \in Y} c(\{x\}, \{y_i\})$,
\item $c_a(\emptyset, \{y\}) = \kappa^+ > max_{x_i \in X} c(\{x_i\}, \{y\})$,
\item $c_a(A, \{y\})=\kappa > max_{x_i \in A} c(\{x_i\}, \{y\})$ for $|A|>1$, and 
\item $c_a(\{x\},B)=\kappa > max_{y_i \in B} c(\{x\}, \{y_i\})$ for $|B|>1$.
\item $c_a(A,B)=0$ for $|A|>1$ and $|B|>1$ to satisfy Proposition~\ref{prop:optimal.assignment.equal.probs.max.plus.transform}.
  Here we assume $\kappa^+ > \kappa$. 
\end{itemize}

This structure, depicted in Table~\ref{tab:cM.c.infty.max}, avoids any assignment from a singleton to a larger set as the cost becomes $\kappa$, and, thus, prioritizes the assignment between singletons, even in the case that $\mu$ and $\nu$ are not probabilities.

This type of assignment can be represented as a graph with two components (i.e., two sets of nodes, and then the nodes in a set cannot be reached from nodes in the other set). One set corresponds to the singletons and the other to the non-singletons. When $|X|=|Y|$ we can define cost functions where this property holds for each cardinality (or for some cardinalities). In this case, we can also represent the structure in terms of graphs where we have a component for each cardinality.

The cost among non-singletons is zero so that the two assignments $a$ and $assg$ have the same overall costs. Other costs different to zero will provide the same assignments $a$ and $assg$ but the overall costs may be different. 

\begin{table}[ht]
  \begin{center}
  \small 
  \begin{tabular}{c||c|c|c|}
    non-singletons $|B|>1$ & $\kappa^{+}$  & $\kappa$  & $\dag$  \\
    \hline
    singletons     $\{y\}$&  $\kappa^{+}$    & $c$       & $\kappa$  \\
    \hline
    $\emptyset$    & ---           & $\kappa^{+}$  & $\kappa^{+}$ \\
    \hline
    \hline
    ---            & $\emptyset$   & singletons $\{x\}$ & non-singletons $|A|>1$  \\
  \end{tabular}
    \end{center}
  \caption{Graphical representation of a cost function $c_a$ so that its corresponding optimal assignment $assg$ is compatible with another assignment $a$ for the cost function $c$. That is, $a \prec assg$. \label{tab:cM.c.infty.max}}
\end{table}

Proposition~\ref{prop:optimal.assignment.equal.probs.max.plus.transform} proves that given $c$ and probabilities $\mu$ and $\nu$ we have an equivalent problem $(c_a, \mu, \nu)$ that returns the same assignment for the singletons. We can observe that each of the values $p(x)=\mu(\{x\})$ (same for $q(y)=\nu(\{y\})$) appears a different number of times. So, we may consider extending $c$ into $c_a$ so that the cost of the assignments is similar. In other words, for a pair $(p(x_i),q(y_j))$, the cost of all related assignments in $c_a$ is the same as $c(x_i, y_j)$. For example, if we consider the pair $(p(x_1)=0.2, q(y_3)=0.6)$ we have that this pair appears $4\times 1$ times in Table~\ref{tab:additive.mu.to.emptyset}. This observation allows us to construct a cost function for our optimal transport problem in which the cost associated to this pair is $(p(x_1)=0.2, p(y_3)=0.6)$, which is the same as the one of $c(x_1,x_3)$. 

\begin{proposition}
  \label{prop:same.cost.c.ca}
  Let $\mu$ and $\nu$ be additive measures on $X$ and $Y$ and let $p$ and $q$ be the corresponding probability distributions. Let $c$ be the cost function of the optimal transport problem in the probabilistic setting associated to $p$ and $q$. Then, there exists a cost function $c_a$ such that the cost of each pair of probabilities $(p(x_i), q(y_j))$ equals to $c(x_i,y_j)$.
\end{proposition}

The proof of this proposition is by construction, and the expression for the cost function $c_a$ is given in Equation~\ref{eq:ca.eq.c}

\begin{proof}
  First, note that $\tau_{\mu}(A)=\min_{x \in A} p(x)$ and $\tau_{\nu}(A)=\min_{x \in A} q(x)$ according to Proposition~\ref{prop:probability.distribution}. It is obvious that for singletons $c_{a}(\{x_i\}, \{y_j\})$ should be associated to $c(x_i, y_j)$. Similarly, in general, we need to associate $c_{a}(A, B)$ to the pair $(x,y)$ such that it produces $\tau_{\mu}(A)$ and $\tau_{\nu}(B)$. In other words, we consider $c(x_i, x_j)$ where $x_i = \arg \min_{x_i \in A} p(x_i)$ and $y_j = \arg \min_{y_j \in B} q(y_j)$.

  The association between $c_a$ and $c$ is not, in general, an equality. Some probabilities $p_i$ appear several times in $\tau_{\mu}(A)$. The same applies to $q_i$. In fact, the smallest probability $\min_{x_i} p(x_i)$ appears more often than the second smallest one, which appears more often than the third smallest, and so on.

  Let $s_{\mu}(i)$ and $s_{\nu}(i)$ define a permutation so that
  \begin{equation}
    \label{eq:permutation.si.p}
    p(x_{s_{\mu}(1)}) \leq p(x_{s_{\mu}(2)}) \leq \dots \leq p(x_{s_{\mu}(n)})
  \end{equation}
  \begin{equation}
    \nonumber
    q(x_{s_{\nu}(1)}) \leq q(x_{s_{\nu}(2)}) \leq \dots \leq q(x_{s_{\nu}(m)})
  \end{equation}

  Then, $p(x_{s_{\mu}(n)})$ appears in $\tau_{\mu(A)}$ only in one set (i.e., $\{x_{s_{\mu}(n)}\}$), the second largest value $p(x_{s_{\mu}(n-1)})$ appears in two sets (i.e., $\{x_{s_{\mu}(n-1)}\}$, $\{x_{s_{\mu}(n)}, x_{s_{\mu}(n-1)}\}$). In general, the $i$th largest value appears in $2^{n-1}$ sets. The same applies to the values for $q$. Observe Table~\ref{tab:additive.mu}. For $\tau_{\mu}$ we have that $p(\{x_3\})=0.5$ appears only one, $p(\{x_2\})=0.3$ appears twice, and $p(\{x_1\})=0.2$ appears 4 times. Same applies to $q$ although in that case as $q(\{x_1\})=q(\{x_2\})=0.2$, this is not so easily observed.

  Let $s^{-1}_{\mu}(i)$ be the order of the $i$th variable, according to the permutation in Equation~\ref{eq:permutation.si.p}. Similarly, $s^{-1}_{\nu}(j)$ corresponds to the order of the $j$th variable. Then, if the pair associated to $(A,B)$ is $(x_i, x_j)$, we need to weight the associated cost $c(x_i, x_j)$ considering the fact that $p(x_i)$ probability appears several times. I.e., $p(x_i)$ appears $2^{n-s^{-1}_{\mu}(i)}$ times. Similarly, $q(y_j)$ appears $2^{m-s^{-1}_{\nu}(j)}$ times. Therefore, we define:

  \begin{eqnarray}
    \label{eq:ca.eq.c}
    c_{a}(A, B)
    &=&\frac{1}{2^{n-i'}} \frac{1}{2^{m-j'}} c\left(\arg \min_{x_i \in A} p(x_i),\arg \min_{y_j \in B} q(y_j)\right) \nonumber \\
    &=&\frac{1}{2^{n-s^{-1}_{\mu}(i)}} \frac{1}{2^{m-s^{-1}_{\nu}(j)}} c(x_i, y_j) \nonumber \\
  \end{eqnarray}
  where $i=\arg \min_{x_i \in A} p(x_i)$ and $j=\arg \min_{y_j \in B} q(y_j)$.

  Then, we can observe that the pair $(p(x_{s_{\mu}(n)}), q(y_{s_{\mu}(m)}))$ appears only once in both $\tau_{\mu}$ and $\tau_{\nu}$ and
  $$c_a(\{x_{s_{\mu}(n)}\}, \{y_{s_{\mu}(m)}\}) =c(x_{s_{\mu}(n)}, y_{s_{\mu}(m)})$$

  Then, for a pair $\{x_i\},\{y_j\}$ we have that $p(x_i)$ will appear $2^{n-s^{-1}_{\mu}(i)}$ times and $q(y_j)$ will appear $2^{n-s^{-1}_{\nu}(j)}$ times. I.e., there will be $2^{n-s^{-1}_{\mu}(i)} 2^{m-s^{-1}_{\nu}(j)}$ pairs of sets $(A,B)$ with $c_a$ associated to the same $c$. 
\end{proof}

%\begin{tabular}{ L | $l$ | $l$ || $l$ $l$ $l$ | $l$ $l$ $l$ | $l$ |}

If $\mu$ and $\nu$ are additive measures; then, if we define $c_a$ according to Equation~\ref{eq:ca.eq.c}, and we compute the solutions of the original problem (i.e., $a$) and of our extension (i.e., $assg$), then they are such that $a \prec assg$. In addition the two assignments have the same cost. 

Nevertheless, this result is about building the cost function $c_a$ for non-singletons taking into account the order of the values for the singletons. That is, using the ordering or permutation of elements $s_{\mu}(1), \dots, s{\mu}(n)$, as well as $s_{\nu}(1), \dots, s{\nu}(n)$. That is, the cost function $c_a$ depends on both $\mu$ and $\nu$.

The next definition defines this cost function explicitly. It follows Equation~\ref{eq:ca.eq.c}. We write $c_a(A,B;\mu,\nu)$ to make this dependence on $\mu$ and $\nu$ explicit. Note also that this definition is for arbitrary fuzzy measures and not only for additive ones. Then, we provide the proposition that establishes the consistency of the assignments. 

\begin{definition}
  \label{def:ca.c.same.cost}
  Let $\mu$ and $\nu$ non-additive measures on $X$ and $Y$. Let $\kappa$ an arbitrary value such that $\kappa > \max c(x,y)$. Then, we define the cost function $c_a: 2^X \times 2^Y \rightarrow {\mathbb R}^+$ as follows
  $$c_a(A,B; \mu, \nu) = \left\{
  \begin{array}{cc}
    0                                                           & A=\emptyset, B=\emptyset \\
    \kappa                                                      & A=\emptyset, B \neq \emptyset \\
    \kappa                                                      & A \neq \emptyset, B=\emptyset \\
    \frac{1}{2^{n-s^{-1}_{\mu}(i)}} \frac{1}{2^{m-s^{-1}_{\nu}(j)}} c(x_i, y_j) & otherwise
  \end{array}
  \right.
  $$
\end{definition}

\begin{proposition}
  Let $\mu$ and $\nu$ additive measures on $X$ and $Y$. Let $c$ a cost function on $X \times Y$, and let $c_a$ the cost function according to Definition~\ref{def:ca.c.same.cost}. Let $a$ be the solution of the optimal transport problem for $c$, and let $assg$ be the solution of the optimal transport problem for $c_a$. Then, $a$ and $assg$ are such that $a \prec assg$.
\end{proposition}

\begin{proof}
  This is a consequence of Proposition~\ref{prop:same.cost.c.ca} as the cost of each pair in $c_a$ is the same as in the cost function $c$. In addition, as $c_a(\emptyset,B)=c_a(A,\emptyset)>c(\{x\},\{y\})$ for all $x,y$, then there is no assignment to $\emptyset$. 
\end{proof}

\subsection{On the implementation of the optimal transport problem}
We discuss in this section how to find a solution for the optimal transport problem for the three different problems established above. We will start with the case of basic probability assignment.

It can be observed that the problem introduced in Definition~\ref{def:OTp.bpa} can be seen as an assignment of probabilities in $2^{X}$ instead of probabilities on $X$. Therefore, the same techniques and approaches used for classical optimal transport problems can be applied here. 

In contrast, the problem stated in Equation~\ref{eq:OTp.mobius} needs a different approach. We can observe that it has some similarities with the one for basic probability assignment but in this case the values of the M\"obius transform can be negative. In addition, the objective function includes an absolute value function. An objective function with an absolute value of a linear expression can be transformed into an equivalent one without the absolute value by means of introducing a new variable. The problem will still be linear. See e.g.~\cite{ref:Granger.Yu.Zhou.2022}. More particularly, if we have an objective function of this form
$$\min OF+|t|$$
we will rewrite it as
$$\min OF+t'$$
and add two additional constraints, $+t \leq t'$ and $-t \leq t'$. This process requires quite a few additional constraints (two for each variable). As there are $2^{|X|} \times 2^{|X|}$ variables, this corresponds to $2\cdot 2^{2|X|}$ additional constraints.

The optimal transport problem associated to the $(max, +)$-transform has a form similar to the one for basic probability assignments. The assignment needs to be positive. Therefore, the problem is also a linear optimization problem with linear constraints. We have developed software in Python for the computation of this optimal problem. Software will be made available~\cite{ref:mdai.code}.

\subsection{Wasserstein distance}
The definition of the optimal transport problem for non-additive measures permits us to consider the definition of distances based on this problem. More particularly, we can consider the definition of a Wasserstein-like discrepancy following Equation~\ref{eq:Wasserstein.probs}. Our definition is based on the definition of the transport problem based on the $(\max, +)$-transform.

\begin{definition}
  \label{def:Wasserstein.nam}
  Let $\mu$ and $\nu$ be non-additive measures on $X$, with $(max, +)$-transforms $\tau_{\mu}$ and $\tau_{\nu}$, respectively. Let $c_a$ a cost function. Let $\Pi(\tau_{\mu},\tau_{\nu})$ be the set of all assignments that are compatible with $\tau_{\mu}$ and $\tau_{\nu}$. Then, we define the Wasserstein-like discrepancy for $\mu$ and $\nu$ given $c_a$ as 
  \begin{equation}
  \label{eq:Wasserstein.nam}
  d_{c_a}(\mu,\nu)=\inf_{assg \in \Pi(\tau_{\mu},\tau_{\nu})} \sum_{A \subseteq X}\sum_{B \subseteq X} c_{a}(A,B) assg(A,B)
  \end{equation}
\end{definition}

It is clear that when $\mu=\nu$ we have that $d_{c_a}(\mu,\nu)$ is zero, and that if there a single optimal assignment, the expression is symmetric. Other properties of this definition need to be further studied. In particular, the triangle inequality so that the discrepancy is, in fact, a distance. Naturally, some properties will depend on the cost function, and on the type of solution the optimization problem gives as result. 

\section{Conclusions and future work}
In this paper we have studied the transport problem for non-additive measures. We have discussed the difficulties of this problem. Then, we have proposed three different definitions, one for belief functions and basic probability assignments and two others for arbitrary non-additive measures. The latter are based on the M\"obius transform and the $(max, +)$-transform. Different formulations stress different ways of looking to the measures. They deal differently with positive and negative interactions. The M\"obius transform can be negative, but not the $(max, +)$-transform. They also deal differently with the fact that the total mass of a non-additive measure can be seen as different to one. 

We have proven some properties, including the ones that establish that our definitions are proper extensions of standard optimal transport problems. We have also briefly discussed the complexity of the related optimization problems.

In the paper we have discussed the cost functions that we need to define on subsets of $X$ and $Y$ and some relationships with the ones for probability distributions. What kind of cost functions are relevant and meaningful for non-additive measures requires is a research direction. We have also provided a discussion on how the problems can be solved numerically. All the problems stated here are linear optimization problems with linear constraints. Nevertheless, the assignment is a function from $2^{|X|} \times 2^{|Y|}$ that makes the problem costly for not so large $X$ and $Y$. Therefore another direction is to work on computational efficient solutions, some may depend on the type of cost functions used. 

\section{Acknowledgments}
This work was partially supported by the Wallenberg AI, Autonomous Systems and Software Program (WASP) funded by the Knut and Alice Wallenberg Foundation


\begin{thebibliography}{99}


 
\bibitem{ref:Agahi.2020}
Agahi, H. (2020) A generalized Hellinger distance for Choquet integral, Fuzzy Sets and Systems 396 42-50. 

\bibitem{ref:Agahi.Yadollahzadeh.2021}
  Agahi, H., Yadollahzadeh, M. (2021) On f-divergence for $\sigma$-$\oplus$-measures, Soft Computing 25 9781–9787. 
  
\bibitem{ref:Arjovsky.Chintala.Bottou.2017}
Arjovsky, M., Chintala, S., Bottou, L. (2017) Wasserstein Generative Adversarial Networks, ICML 2017, PMLR 214–223.


\bibitem{ref:Bardozzo.et.al.2021}
  Bardozzo, F., de la Osa, B., Horansk\'a, L., Fumanal-Idocin, J., delli Priscoli, M., Troiano, L., Tagliaferri, R., Fern\'andez, J., Bustince, H. (2021) Sugeno integral generalization applied to improve adaptive image binarization, Inf. Fusion 68 37-45. 

  \bibitem{ref:Beliakov.2022}
  Beliakov, B. (2022) Knapsack problems with dependencies through non-additive measures and Choquet integral, European Journal of Operational Research 301:1 277-286. 

Author links open overlay panel

\bibitem{ref:Benvenuti.Mesiar.Vivona.2002}
Benvenuti, P., Mesiar, R., Vivona, D. (2002) Monotone set functions-based integrals, in E. Pap (ed.) Handbook of Measure Theory, North-Holland, 1329-1379. 

\bibitem{ref:Bogachev.Kolesnikov.2012}
  Bogachev, V. I., Kolesnikov, A. V. (2012) The Monge-Kantorovich problem: achievements, connections, and perspectives, Russ. Math. Surv. 67:5 1-110.


\bibitem{ref:Bronevich.Rozenberg.2021}
Bronevich, A. G., Rozenberg, I. N. (2021) The measurement of relations on belief functions based on the Kantorovich problem and the Wasserstein metric, Int. J. of Approx. Reasoning 131 108-135.

\bibitem{ref:Chateauneuf.1996}
Chateauneuf, A. (1996) Decomposable measures, distorted probabilities and concave capacities, Mathematical Social Sciences 31 19-37. 
%% Results on distorted probabilities. Cited in the chapter on DP

\bibitem{ref:Chateauneuf.Jaffray.1989}
Chateauneuf, A., Jaffray, J.-Y. (1989) Some characterizations of lower probabilities and other monotone capacities through the use of M\"obius inversion, Mathematical Social Sciences 17:3 263-283. 


\bibitem{ref:Choquet.1953.54} 
Choquet, G. (1953/54) Theory of capacities, Ann. Inst. Fourier 5 131-295. 

\bibitem{ref:Denneberg.1994}
Denneberg, D. (1994) Non Additive Measure and Integral, Kluwer Academic Publishers. 


\bibitem{ref:Edwards.1953}
  Edwards, W. (1953) Probability-preferences in gambling, American Journal of Psychology 66 349-364.

\bibitem{ref:Gal.Niculescu.2019}
  Gal, S. G., Niculescu, C. P. (2019) Kantorovich's mass transport problem for capacities, arXiv: 1907.03749v4. 


\bibitem{ref:Gilboa.2009} 
Gilboa, I. (2009) Theory of decision under uncertainty, Cambridge University Press. 


\bibitem{ref:Granger.Yu.Zhou.2022}
Granger, B., Yu, M., Zhou, K. (2022) Optimization with absolute values,
\url{https://optimization.mccormick.northwestern.edu/index.php/Optimization_with_absolute_values}
(accessed 9 March 2022)

\bibitem{ref:Honda.2014}
Honda, A. (2014) Entropy of capacity, in V. Torra, Y. Narukawa, M. Sugeno (eds.) Non-additive measures, Springer 79-95. 
\bibitem{ref:Honda.Nakano.Oakazaki.2002:Distortion}
Honda, A., Nakano, T., Okazaki, Y. (2002) Distortion of fuzzy measures, Proc. of the SCIS/ISIS conference. 
%% Cited in the chapter on fuzzy measures, for distorted probabilities. This is the paper that includes the comparison between the number of DP and of general fuzzy measures. 

\bibitem{ref:Honda.2002:Subjective}
Honda, A., Nakano, T., Okazaki, Y. (2002) Subjective evaluation based on distorted probability, Proc. of the SCIS/ISIS conference. 
%% Cited in the chapter on fuzzy measures, for distorted probabilities. 

\bibitem{ref:Kantorovich.1942}
  Kantorovich, L. V. (1942) On mass moving, Dokl. Akad. Nauk SSSR 37 (7-8) 227-229.
  
\bibitem{ref:Marco-Detchart.et.al.2021}
  Marco-Detchart, C., Lucca, G., Lopez-Molina, C., De Miguel, L., Pereira Dimuro, G., Bustince, H. (2021) Neuro-inspired edge feature fusion using Choquet integrals, Inf. Sci. 581 740-754


\bibitem{ref:Mesiar.1999}
Mesiar, R. (1999) Generalizations of $k$-order additive discrete fuzzy measures, Fuzzy Sets and Systems 102 423-428. 

\bibitem{ref:Mesiar.et.al.2016}
  Mesiar, R., Koles\'arov\'a, A., Bustince, H., Dimuro, G. P., Bedregal, B. C. (2016) Fusion functions based discrete Choquet-like integrals, European Journal of Operational Research 252:2 601-609.Decision Support


\bibitem{ref:Mesiar.Mesiarova.2004}
Mesiar, R., Mesiarov\'a, A. (2004) Fuzzy integrals, MDAI 2004, Lecture Notes in Artificial Intelligence 3131 7-14. 


\bibitem{ref:Pereira.Figueira.Marques.2020}
  Pereira, M. A., Figueira, J. R., Marques, R. C. (2020) Using a Choquet integral-based approach for incorporating decision-maker’s preference judgments in a Data Envelopment Analysis model, European Journal of Operational Research 284:3 1016-1030.

\bibitem{ref:Pereira-Dimuro.et.al.2020:CI.generalizations}
  Pereira Dimuro, G., Fern\'andez, J., Bedregal, B. R. C., Mesiar, R., Sanz, J. A., Lucca, G., Bustince, H. (2020) The state-of-art of the generalizations of the Choquet integral: From aggregation and pre-aggregation to ordered directionally monotone functions, Inf. Fusion 57 27-43

  \bibitem{ref:Pereira-Dimuro.et.al.2020}
  Pereira Dimuro, G., Lucca, G., Bedregal, B. R. C., Mesiar, R., Sanz, J. A., Lin, C.-T., Bustince, H. (2020) Generalized CF1F2-integrals: From Choquet-like aggregation to ordered directionally monotone functions, Fuzzy Sets Systems 378 44-67. 

\bibitem{ref:Santambrogio.2015}
Santambrogio, F. (2015) Optimal Transport for Applied Mathematicians - Calculus of Variations, PDEs and Modeling, Birkh\"auser. 


\bibitem{ref:Sugeno.1972}
Sugeno, M. (1972) Fuzzy measures and fuzzy integrals (in Japanese), Trans. of the Soc. of Instrument and Control Engineers 8:2 

\bibitem{ref:Sugeno.1974}
Sugeno, M. (1974) Theory of Fuzzy Integrals and its Applications, Ph. D. Dissertation, Tokyo Institute of Technology, Tokyo, Japan. 

\bibitem{ref:Tehrani.et.al.2012}
Tehrani, A. F., Cheng, W., Dembczynski, K., H\"ullermeier, E. (2012) Learning monotone nonlinear models using the Choquet integral, Machine Learning 89 183-211. 

\bibitem{ref:Torra.2017:entropy} 
Torra, V. (2017) Entropy for non-additive measures in continuous domains, Fuzzy sets and systems, Fuzzy sets and systems 324 49-59. 

\bibitem{ref:Torra.Guillen.Santolino.2018:INFUS}
Torra, V., Guillen, M., Santolino, M. (2018) Continuous m-dimensional distorted probabilities, Information Fusion 44 97-102. 

\bibitem{ref:Torra.Narukawa.Sugeno.2013}
  Torra, V., Narukawa, Y., Sugeno, M. (eds.) (2013) Non-additive measures: theory and applications, Springer.

\bibitem{ref:Torra.Narukawa.Sugeno.2016} 
  Torra, V., Narukawa, Y., Sugeno, M. (2016) On the $f$-divergence for non-additive measures, Fuzzy sets and systems 292 364-379.

\bibitem{ref:Torra.Narukawa.Sugeno.2020}  
Torra, V., Narukawa, Y., Sugeno, M. (2020) On the f-divergence for discrete non-additive measures, Inf. Sci. 512 50-63.

\bibitem{ref:Torra.Narukawa.Abril.2014}
Torra, V., Narukawa, Y., Abril, D. (2014) Comparing fuzzy measures through their Möbius transform, Proc. 17th Int. Conf. Information Fusion (FUSION). 

\bibitem{ref:Torra.2023:max.plus.additive}
  Torra, V. (2022) $(Max,\oplus)$-transforms and genetic algorithms for fuzzy measure identification, Fuzzy sets and systems 451 253-265. \url{https://doi.org/10.1016/j.fss.2022.09.008}

\bibitem{ref:Torra.Narukawa.Sugeno.Carlson.2013}
  Torra, V., Narukawa, Y., Sugeno, M., Carlson, M. (2013) Hellinger distance for fuzzy measures, EUSFLAT 2013.

\bibitem{ref:Villani.2003}
Villani, C. (2003) Topics in optimal transportation, AMS. 

\bibitem{ref:Villani.2008}
  Villani, C. (2008) Optimal Transport: Old and New, Springer.

\bibitem{ref:mdai.code}
  \url{http://www.mdai.cat/code}

\end{thebibliography}
\end{document}